\documentclass[sigconf]{acmart}

\AtBeginDocument{%
  \providecommand\BibTeX{{%
    \normalfont B\kern-0.5em{\scshape i\kern-0.25em b}\kern-0.8em\TeX}}}

\setcopyright{acmcopyright}
\copyrightyear{2018}
\acmYear{2018}
\acmDOI{10.1145/1122445.1122456}

\acmConference[Woodstock '18]{Woodstock '18: ACM Symposium on Neural
  Gaze Detection}{June 03--05, 2018}{Woodstock, NY}
\acmBooktitle{Woodstock '18: ACM Symposium on Neural Gaze Detection,
  June 03--05, 2018, Woodstock, NY}
\acmPrice{15.00}
\acmISBN{978-1-4503-9999-9/18/06}

\begin{document}

\settopmatter{printacmref=false} 
\renewcommand\footnotetextcopyrightpermission[1]{} 
\pagestyle{plain} 
\setlength{\emergencystretch}{4pt}
\title{Selection-based Question Answering of an MOOC}

\author{Atul Sahay}

\affiliation{%
  \institution{Department of Computer Science and Engineering, IIT Bombay}
  \streetaddress{Powai}
  \city{Mumbai}
  \state{Maharashtra}
  \country{India}}
  \email{atulsahay@cse.iitb.ac.in}

\author{Smita Gholkar}
\affiliation{%
  \institution{Department of Computer Science and Engineering, IIT Bombay}
  \streetaddress{Powai}
  \city{Mumbai}
  \country{Maharashtra}
  \country{India}}
\email{smitagh@cse.iitb.ac.in}

\author{Kavi Arya}
\affiliation{%
 \institution{Department of Computer Science and Engineering, IIT Bombay}
 \streetaddress{Powai}
 \city{Mumbai}
 \state{Maharashtra}
 \country{India}}
 \email{kavi@cse.iitb.ac.in}


\begin{abstract}
e-Yantra Robotics Competition (eYRC) is a unique Robotics Competition hosted by IIT Bombay that is actually an Embedded Systems and Robotics MOOC. Registrations have been growing exponentially in  each year from 4500 in 2012 to over 34000 in 2019. In this 5 month long competition students learn complex skills under severe time pressure and have access to a discussion forum to post doubts about the  learning material. Responding to questions in real-time is a challenge for project staff. Here, we illustrate the advantage of Deep Learning for real-time question answering in the eYRC discussion forum. We illustrate the advantage of Transformer based contextual embedding mechanisms such as  Bidirectional Encoder Representation From Transformer (BERT) over word embedding mechanisms such as Word2Vec. We propose a weighted similarity metric as a measure of matching and find it more reliable than Content-Content or Title-Title similarities alone. The automation of replying to questions has brought the turn around response time(TART) down from a minimum of 21 mins to a minimum of 0.3 secs. 
\end{abstract}

\ccsdesc[500]{Computing methodologies~Information extraction}
\ccsdesc[300]{Computing methodologies~Ranking}
\ccsdesc[300]{Computing methodologies~Neural networks}
\ccsdesc[100]{Computing methodologies~Ontology engineering}

\keywords{Virtual Assistant, Natural Language Programming, Deep Learning, BERT, Word2Vec, Context Similarity, Text Analytics}


\maketitle

\section{Introduction}
e-Yantra Robotics Competition (eYRC) is a pan-India robotics outreach initiative hosted at IIT Bombay where  undergraduate students from Engineering, Polytechnic and Science colleges participate in vast numbers. Students teams of 4 and submit tasks as part of team work. Teams are evaluated and scored on their performance to progress further in the competition. This year, however, the paradigm of eYRC has changed to be more inclusive and includes all participating students in the eYRC fold for the first half (Stage-1) of the competition. This implies over 34000 students will engage with e-Yantra staff through a discussion forum posing questions about the learning material in real time. It is imperative to develop a process where e-Yantra can engage with students 24x7 during the competition to reduce participant stress  and optimise e-Yantra resources.  We need automation to help us cope with  questions posed by  participants during the competition.

  The competition has a Project Based Learning (PBL) format i.e.Learning by Competing and Competing by Learning \cite{learning}, where students work  with minimal personal contact and all assistance online. The goal is to help  students learn  challenging concepts such as microcontroller programming, Robotic Operating Systems (ROS), Image processing, control system design, etc. - remotely with online guidance from e-Yantra. The training develops the ability to apply concepts in practical situations. To assist the process e-Yantra engages a discussion forum. On the forum, students post questions where they ask e-Yantra staff for clarification and pointers to resources to learn from. As tasks get time critical, there is a flood of questions on the discussion forum relating to the task in hand or concepts they need to learn. Many questions are repeated as students fail to search for a similar issue posed previously on the discussion forum (by similar here we mean in context and content). With the abundance of algorithms, software libraries, hardware infrastructure and knowledge in the domain of Natural Language Processing for Education setting \cite{learning2}, the use of technology to support e-Yantra's staff by automating answers to repeated questions is natural. We found section 3.A of the survey reported in literature \cite{otter2018survey} and "mining social media" and "machine reading" sections of the survey \cite{hirschberg2015advances} useful and relevant to our problem. 

Most questions posed on the discussion forum of eYRC are similar in context and content but not syntax. Thus, these questions don't show up in the inbuilt "Word" only based search provided by the discussion forum. The purpose of our work carried is to - 1. Provide a response in less than a minimum of 21 mins of student questions to 0.3 seconds; 2. Reduce the cognitive load of students in searching for previously answered questions similar to theirs on the discussion forum; 3. Reduce overheads for e-Yantra members answering similar questions repeatedly; and 4. Automate the process for answering repeated questions given the sheer volume of participants ( $\sim$ 34340 this year - 2019).

In Section 2, we illustrate our architecture's model; in Section 3, we elaborate on our approach as an algorithmic process; further in Section 4, we discuss the setup and observations and finally conclude in Section 5.

\section{Description of Basic Idea and Model Architecture}
\begin{figure}[h]
\centerline{\includegraphics[width=8cm, height=4cm]{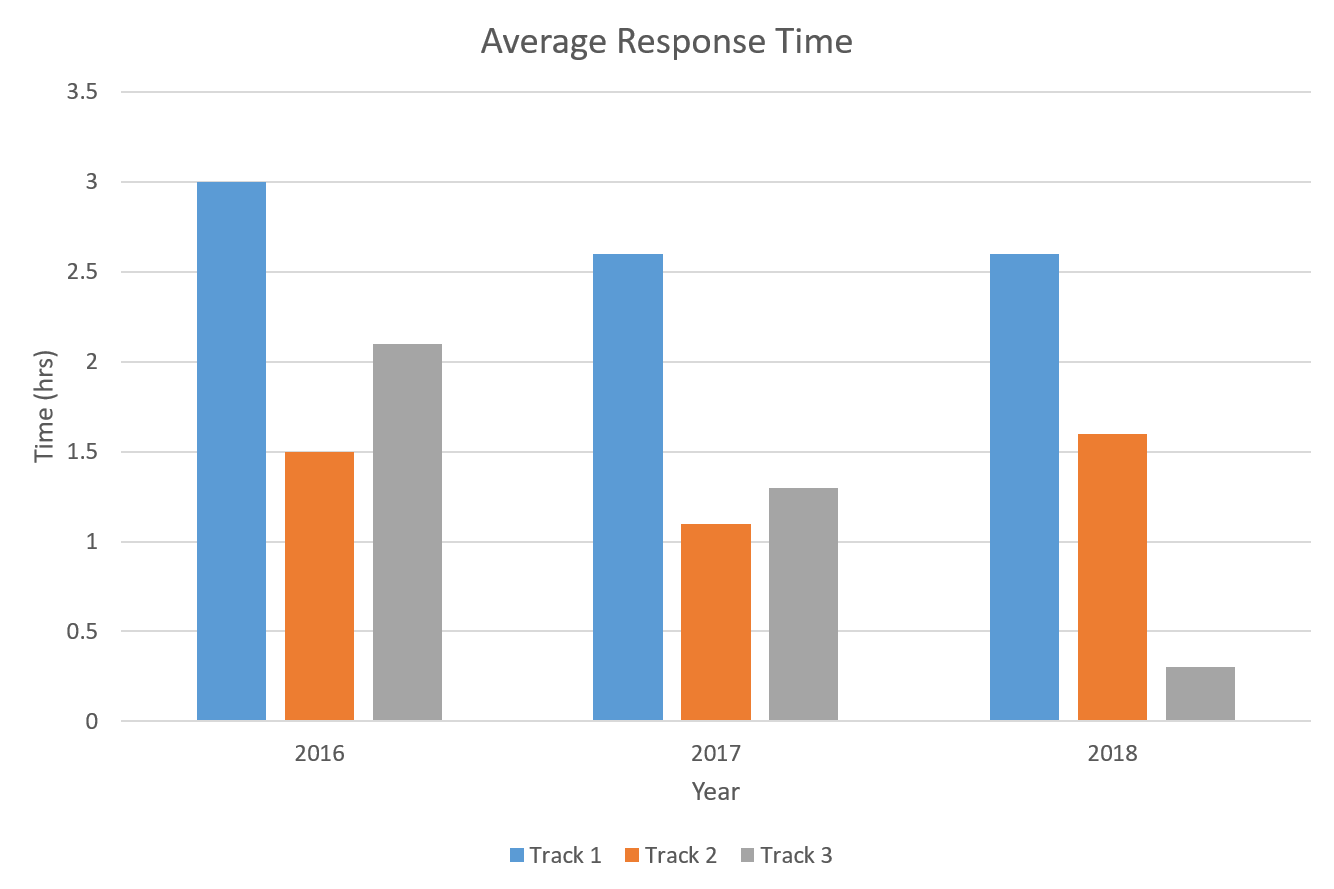}}
\caption{Average Response times for the last three years on discussion forum}
\label{figure1}
\end{figure}

\begin{figure}[h]
\centerline{\includegraphics[width=8cm, height=4cm]{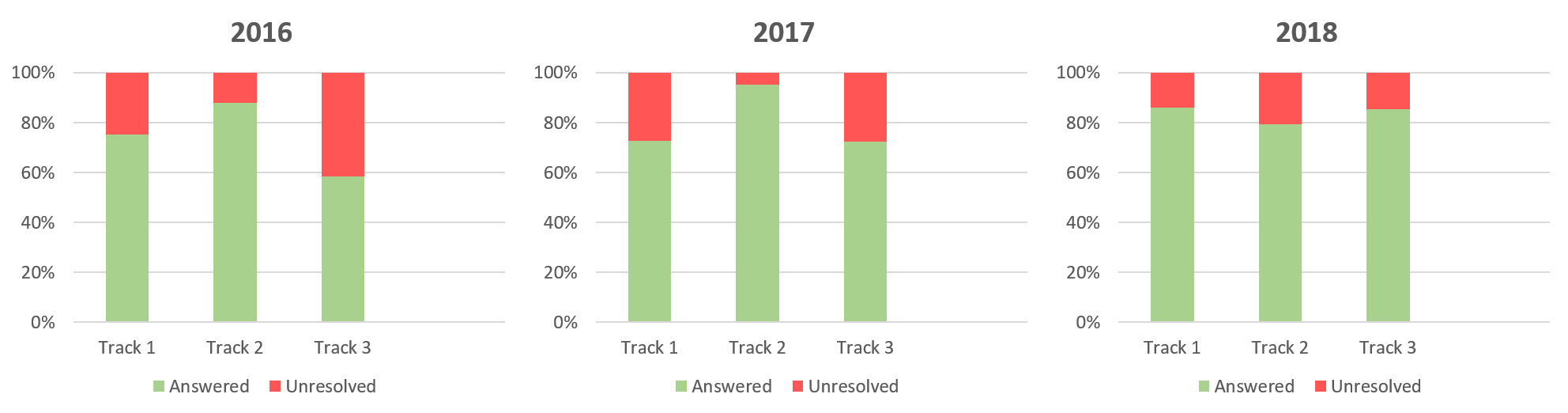}}
\caption{Numbers of unresolved queries per track across three years}
\label{figure2}
\end{figure}

The primary purpose of our work is to reduce the response time for answering a question on e-Yantra's discussion forum for the duration of eYRC. The lowest recorded response time has been 21 mins in the year 2018 '' Fig. \ref{figure1}'' in a theme track whose unresolved questions consisted of 18-20\% of total questions asked ''Fig. \ref{figure2}''. In order to resolve all questions, it is important to differentiate between new questions, thread follow ups and repeated questions. Repeated questions can be answered in an automated manner if the instructor has provided an answer to such a question earlier. Else it will be useful to flag it as an unresolved question. This serves the dual purpose of not just reducing average response time to make it closer to real-time answering; but also to reduce the number of unresolved questions. In order to automate the answering of repeat questions, our automation system needs to understand the perspective and objective of a participant's question \cite{learning3} \cite{ijariit}.

\begin{figure}[h]
\centerline{\includegraphics[width=8cm, height=4cm]{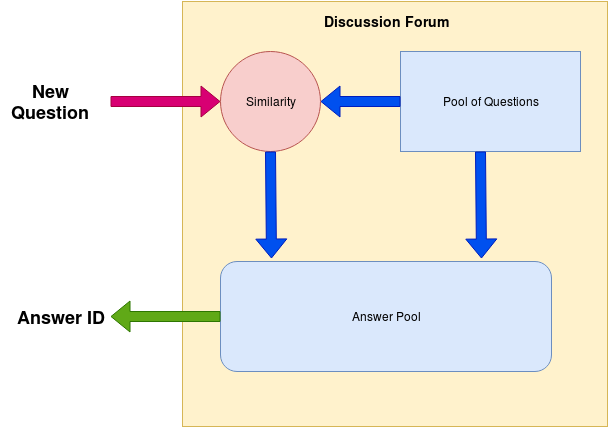}}
\caption{Model Architecture}
\label{figure3}
\end{figure}

The model architecture is as shown in ''Fig. \ref{figure3}''. Say student 'x' asks a question at  time instance 't'. All questions asked from the time interval [0,t] by students including 'x' are considered as the knowledge base from whose data a response has  to be discovered. The individual question at instance 't' by student 'x' is treated as a "New Question". The similarity of this new question is computed against a knowledge base or "Pool of Questions." The top five similarity matches are suggested as possible answers if a question context similarity is obtained with above 70\% similarity from the corresponding answers of the Pool of Questions pointed to by the matching query ids. The top five answers are suggested in order to broaden the possible solutions to the student's question; rather than just an exact matching answer.

\section{Our Approach}
We followed a 3-step architecture described below to match a New Question at any time instance with its contextually closest candidate found in the Knowledge base. The top five candidates have a unique query id attached to it. With this query id information in hand, we retrieve all the hierarchical discussions happened with that query in the past. Therefore, in our model we try to find semantically similar doubts already in the Knowledge base with a confidence score higher than a predefined baseline(in our case around 70\%) to answer a query. Throughout our approach, we strictly follow a term-based matching techniques instead of  character based matching as mentioned in the survey report \cite{gomaa2013survey}.

\subsection{Step 1: Building the Knowledge Base}
In order to comply with the model architecture, we build a Knowledge base that comprises tag sets, titles, summaries of queries and a unique id given by the discussion forum's book keeping mechanism. We collected over 13045  questions across 3 years of discussions (2016-2018) of e-Yantra's Robotics competitions held to date. After cleaning this data we were left with over 10000 questions in our Knowledge base.
\begin{table}[h]
\centering
\begin{tabular}{|p{10mm}|p{10mm}|p{18mm}|p{18mm}|}
\hline
QUERY ID & TITLE & MAIN CONTENT & TAGS \\ \hline
je32511i & Unable to see demo video & we did not watch the demo video and now its not there in the portal . Please help!!! & transporter\_bot
 \\ \hline
je0td4d1 & Float Division Error & Sir we are trying to find center of color marker using moments, but continuously we are getting this "FLOAT DIVISION BY ZERO" error, we are not getting problem. Help us as soon as possible. Thank You in advance.   & planter\_bot
\\ \hline
jdbjt4ko & blender problem & When we are trying to move robot through loc send by xbee in blender blender stops responding     & transporter\_bot
 \\ \hline
\end{tabular}
\label{table1}
\caption{An illustration of our Knowledge Base}
\end{table}
\subsection{Step 2: Sentence Embedding}
After building the Knowledge base, we need to encode each item in it; this encoding is called an embedding. We focus on two embedding strategies - word2vec and BERT. The embedding has to be performed on a New Question as well and matched to find similar questions. Here we discuss the process of each embedding strategy.

\subsubsection{Word2Vec: }
A shallow 2 layer neural architecture, used to produce the word embedding \cite{word2vec}. These models are trained on a large corpus of text as the questions often contain paragraphs made up of a number of sentences. We made use of spacy's ( open-source software library for NLP tasks) word2vec model that provides a 300 dimensional vector for words in the vocabulary and a <OOV> tag for the out-of-vocabulary or unseen words.

The Sentence embedding is then the orderly concatenation of all such word vectors as defined by the spacy library.

\subsubsection{BERT Encoder: }  BERT( Bidirectional Encoder Representation From Transformer)\cite{BERT} is a transformer\cite{TRANSFORMER} based architecture pretrained on the large language representations that can be used to extract contextual features from the text or can be fine tuned for the specific tasks of classification or question-answering. It is a model that provides a 768 dimensional vector for a sentence.

BERT tackles two inefficiencies of the word2vec model, first embeddings constructed by this model is useful for keyword / search expansion, meaning-based search that is the whole purpose of using it beforehand, as we match student's question contexts. We want to accurately retrieve  contextual meanings, even if there are no keywords or phrase overlaps.

Secondly, and perhaps more importantly, the embeddings constructed by this model captures high quality language features dynamically informed by words around, instead of a fixed word representation by word2vec that provides the same word representation to the contextually different polysemous words. For example, given two sentences: 

"The cashier in a bank should be a graduate level apprentice." 

"Calcium deposits are found on the banks of the river."

The word2vec model produces the same fixed word representational vector for the word "bank" used in both sentences, while BERT produces a completely different sentence (word) representations. Furthermore, BERT also tackles the out-of-vocabulary inefficiency of word2vec. This disambiguation of word sense \cite{jurafsky} power is inherent to BERT due to its Transformer based sentence embedding where self attention is used over the sentence.   \cite{bert_wsd}. Further the capability of BERT to work with the unstructured and varying rawness of the data makes it versatile and robust over its usage.\cite{Das_2017}.

\subsection{Step 3: Similarity Metrics} 
Each of the query question's word representations either in the form of an individual word set or in the form of a vector in some dimensional space is matched against possible candidates from the Knowledge base for retrieving the contextually or intent similar previously asked question. For this we evaluated a number of similarity measures, described below.

\subsubsection{JACCARD Similarity: } 
Jaccard similarity as a distance measure, depicts how dissimilar two populations are on the basis of overlapping statistics. The range is from [0-1]; with higher values representing more similarity between two populations.

\begin{equation}
    J_{sim}(A,B) = \frac{|A \cap B|}{|A\cup B|}
\end{equation}
Where A and B are the two token sets.

We made use of the individual words of a query as token set and their matches with the Knowledge base.

\subsubsection{Cosine Similarity: } 
Cosine similarity measures the similarity of two non-zero vectors of an inner product space by measuring the cosine of the angle between them. The range of which is from [0-1]; with higher values representing greater similarity between two vectors.

\begin{equation}
    C_{sim}(A,B) = cos(\Theta ) = \frac{A \cdot  B}{\left \| A \right \|\left \| B \right \|}
\end{equation}

Where A and B are the two vectors representing a query and a candidate from the Knowledge base.

\subsubsection{Weighted Similarity: }  We found that the Title of a question conveys subtle information about its content. In our model, we prioritise more on question Title than Main Content. The Title is a gist and gives the precise topic on which a question is based eg. blender problem as depicted in 3rd row of Table 1. Main Content on the other hand is verbose and can have many sentences unrelated to the topic at hand in the larger scheme of steps to build a case for a question. As expected, giving higher weights to the Title's match as compared to the Main Content's match showed good performance.

As a special case of weighted similarity, we normalized the similarity metric ''Eqs. \ref{eq3} - \ref{eq6} '' of the match to obtain the top five matched question similar to the query and observed better results in terms of ranking within the top five.

$$
    title_{query} = BERT(TITLE_{query})  \\
$$
$$
    title_{candidate} = BERT(TITLE_{candidate}) \\
$$
$$
    content_{query} = BERT(CONTENT_{query}) \\
$$    
$$
    content_{candidate} = BERT(CONTENT_{candidate}) \\
$$
\begin{equation}
    t_{sim} = C_{sim}(title_{query}, title_{candidate})
    \label{eq3}
\end{equation}
\begin{equation}
    h_{sim} = C_{sim}(title_{query}, content_{candidate})
    \label{eq4}
\end{equation}
\begin{equation}
    c_{sim} = C_{sim}(content_{query}, content_{candidate})
    \label{eq5}
\end{equation}
\begin{equation}
N_{sim} = \frac{p*t_{sim} + q*h_{sim} + r*c_{sim}}{(p+q+r)}
\label{eq6}
\end{equation}
$$
\text{Where p+q+r >= 1} 
$$

Here $title_{query}$, $title_{candidate}$, $content_{query}$ and $content_{candidate}$ are the word embeddings produced by the BERT encoder. $t_{sim}$, $h_{sim}$ and $c_{sim}$ are the title-title cosine similarity, title-content cosine similarity and content-content similarity between query and candidate respectively. p,q,r are the tunable hyper-parameters. Finally normalised similarity is given by $N_{sim}$.

\section{Setup And Discussion}
\subsection{Model Setup}
Our knowledge base was crawled from the discussion forum into text comma separated value (tsv) files. Each file comprised of questions from eYRC 2016-2018 for all themes (here we refer to projects as themes) of all tracks. eYRC typically consists of three tracks with a minimum of one to a maximum of three themes per track. Over 10000 questions made up the Knowledge base.

\begin{figure}[h]\centerline{\includegraphics[width=8cm, height=5cm]{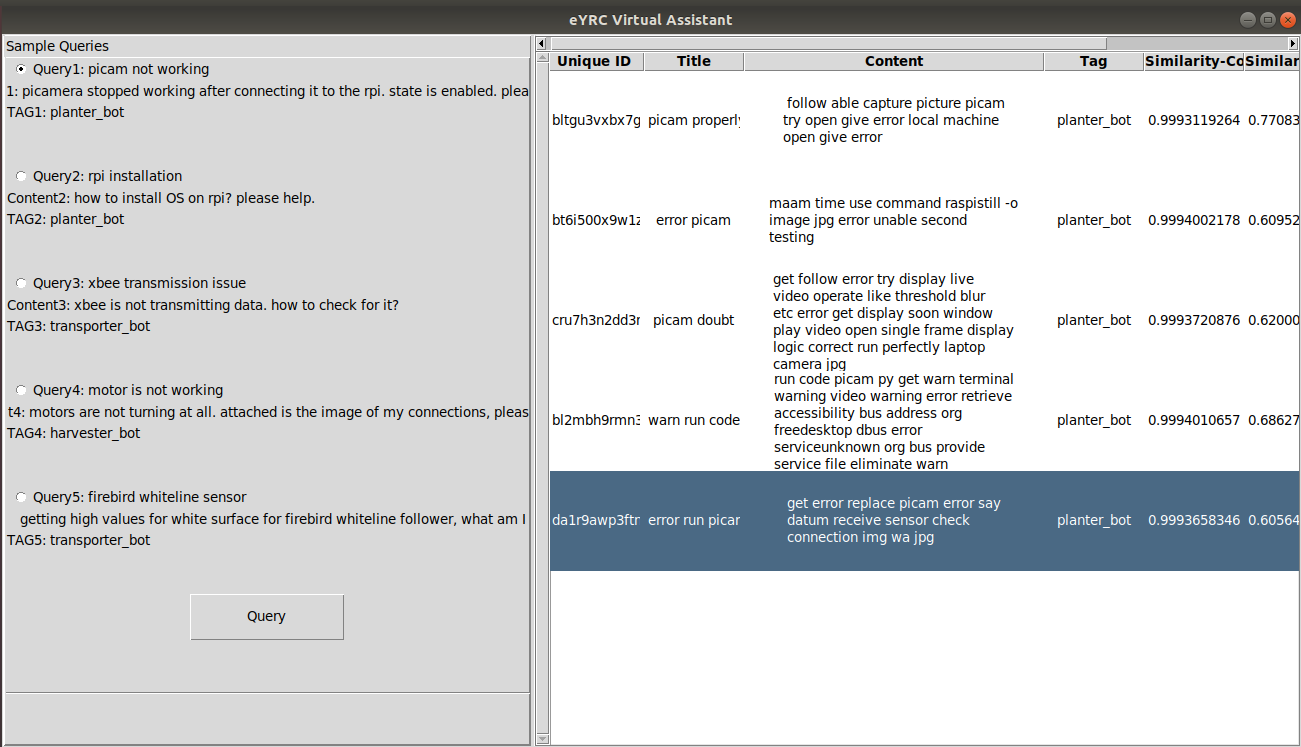}}
\caption{Graphical User Interface for user interaction}
\label{figure4}
\end{figure}

A set of sample Tag, Title and Main Content details are shown in python GUI and calls the code which executes Steps 1-3 in Section 3 on clicking the "Query" button. The screenshot of the GUI and the corresponding query's results are shown in Fig. \ref{figure4}.


\subsection{Discussion}
The performance of sentence based context matching via BERT
gave superior results compared to the word based context matching via Word2Vec as shown in the results and was evaluated for quality of matches manually.

The timings achieved by our automated process vs. the recorded times for response thus far are compared in the table below:

\begin{table}[h]
\centering
\begin{tabular}{|l|l|l|}
\hline
Time Statistics & TART[so far] & TART[Automation] \\ \hline
Minimum & 21 (min)  & 0.32 (secs) \\ \hline
Average & 2.6(hrs) & 152 (secs) \\ \hline
Maximum & 8 (hrs) & 192 (secs)  \\ \hline
\end{tabular}
\caption{Comparison of timings achieved before our automated process vs. after automated process }
\label{table3}
\end{table}

After testing various queries against our Knowledge base, We observe the following - 
\begin{enumerate}
    \item Exact word matches lead to inaccurate results when we want answers to context in the question and hence Jaccard perform undesirably in our context when considered by itself.
    
    \item Even though the numerical value of the cosine similarity is high, its matches in the semantic space are quite misleading with respect to the tag set and objective of the question being asked.
    
    \item Searching for top 5 similar questions and their answer threads retrieved by those unique ids is almost real-time compared to the lowest recorded average response time (here thread refers to the instructors answer and subsequent discussion related to the original question).
    
     \item The minimum TART can be further improved by storing the sentence embeddings obtained from BERT in a database apriori and run a search match on this via querying the database rather than a collection of tsvs as a Knowledge base.
    
    \item Due to the verbosity of Main Content part of questions being matched; even though BERT excelled over word2vec (given word2vec uses a fixed vector space representation for each word irrespective of the dynamism of the context of words around it) in deriving closest results; a normalized weighted similarity metric was used as a match index because the syntax and quantum of Main Content within the questions were of a varied nature.
    
    \item There are themes across tracks which employ same technologies for eg. image processing tasks are assigned based on OpenCV libraries use with python as the scripting language; thus if a query is asked in another TAG set which has been answered, even this id shows up in the list. 
    
    \item We have observed that using Jaccard as the first layer of matching followed by BERT gives more precise context matches. However we need to investigate this further elaborately. A probable explanation of this is that on smaller sentences, like Titles, BERT is not effectively able to model the sentence embeddings as compared to exact word overlap done by Jaccard.
    
\end{enumerate}

\section{Conclusion}
The goal of the  tasks undertaken is to automate answering questions which are repeated, reduce load of the e-Yantra resource and minimize average response time on the discussion forum, and lastly reduce the time to search for an answer for a student participating in eYRC. The task was broken down into a simpler proof of concept demonstrated in this paper and can be scaled up on a server system for making a real-time question answer virtual assistant with application specific to e-Yantra's needs using BERT and weighted similarity metrics.

As demonstrated in table 2. we have achieved considerably lower turn around response time for answering doubts and questions of students given a knowledge base. Going further, we plan to investigate the hypothesis of the Jaccard layer followed by BERT stated in point 8. in the above Section, Save the embeddings in a database to match these encoding and save conversion time whenever a new query comes in the discussion forum, Scale up this model on a server system to be able to cope with the ever growing numbers, The ultimate goal is to build an active learning network to minimize human interference.

\bibliographystyle{ACM-Reference-Format}
\bibliography{sample-base}

\end{document}